\pdfoutput=1
\documentclass[letterpaper, 10 pt, conference]{ieeeconf}
\IEEEoverridecommandlockouts                              %

\title{\LARGE \bf Accelerating db-A$^\textbf{*}$ for Kinodynamic Motion Planning Using Diffusion

}
\author{Julius Franke$^{1}$, Akmaral Moldagalieva\textsuperscript{*}$^{1}$, Pia Hanfeld\textsuperscript{*}$^{1,2}$, and Wolfgang Hönig$^{1}$%
\thanks{$^{1}$ Technical University Berlin, $^{2}$ CASUS, Helmholtz-Zentrum Dresden-Rossendorf, \textsuperscript{*} authors contributed equally.}%
\thanks{Code:\url{https://github.com/juliusfranke/diffusion-motion-planning}.}%
\thanks{The research was partially funded by the Deutsche Forschungsgemeinschaft (DFG, German Research Foundation) - 448549715. Furthermore, it was partially funded by the Center for Advanced Systems Understanding (CASUS), financed by Germany’s Federal Ministry of Education and Research (BMBF), and by the Saxon state government out of the State budget approved by the Saxon State Parliament.
}%
}

\usepackage[bookmarks=true]{hyperref}
\usepackage[nohyperlinks]{acronym}
\usepackage{booktabs}

\usepackage{amssymb}
\usepackage{amsmath}
\usepackage{multirow}
\usepackage[detect-weight=true, detect-family=true]{siunitx}

\usepackage[ruled,vlined,linesnumbered]{algorithm2e}
\SetInd{0.1em}{0.4em}
\SetAlFnt{\footnotesize\sf}
\SetKwRepeat{Do}{do}{while}
\SetKw{KwStep}{step}
\SetKwInput{KwData}{Input}
\SetKwInput{KwResult}{Result}
\SetKwBlock{RepeatForever}{repeat}{}
\SetKw{Continue}{continue}
\SetKwComment{Comment}{$\triangleright$\ }{}
\SetCommentSty{itshape}

\usepackage[capitalise]{cleveref} %
\crefname{equation}{}{} %
\usepackage{flushend}

\usepackage[natbib=true,
    style=ieee,
    citestyle=numeric-comp,
    backend=biber,
    maxbibnames=10,
    maxcitenames=1,
    mincitenames=1,
    doi=false,
    url=false,
    giveninits=true]{biblatex}
\renewcommand*{\bibfont}{\footnotesize}

\usepackage{tikz}
\usetikzlibrary{fit,calc}
\tikzstyle{process} = [rectangle, 
minimum width=1cm, 
minimum height=4cm, 
text centered, 
text width=1cm, 
draw=black, 
fill=orange!30]

\definecolor{forestgreen}{rgb}{0.0, 0.27, 0.13}
\newcommand*{\tikzmk}[1]{\tikz[remember picture,overlay,] \node (#1) {};\ignorespaces}

\newcommand{\markbluelineHalf}[1]{\tikz[remember picture,overlay]{\node[yshift=2pt,xshift=#1,fill=blue!60,opacity=.25,fit={(A)($(A)+(0.55\linewidth,-0.3\baselineskip)$)},rounded corners=4pt] {};}\ignorespaces}

\newcommand{\vu}{\mathbf{u}}    %
\newcommand{\vq}{\mathbf{q}}    %
\newcommand{\mz}{\mathbf{0}}
\newcommand{\mI}{\mathbf{I}}

\newcommand{\seqU}{\mathbf{U}}    %

\newcommand{\seqQ}{\mathbf{Q}}    %

\newcommand{\sW}{\mathcal{W}}
\newcommand{\sM}{\mathcal{M}}
\newcommand{\sU}{\mathcal{U}}   %
\newcommand{\sB}{\mathcal{B}} %
\newcommand{\vf}{\mathbf{f}}    %
\DeclareMathOperator{\step}{step}

\makeatletter
\renewcommand*\env@matrix[1][*\c@MaxMatrixCols c]{%
  \hskip -\arraycolsep
  \let\@ifnextchar\new@ifnextchar
  \array{#1}}
\makeatother

\begin{document}

\maketitle
\thispagestyle{empty}
\pagestyle{empty}

\begin{abstract}

We present a novel approach for generating motion primitives for kinodynamic motion planning using diffusion models. 
The motions generated by our approach are adapted to each problem instance by utilizing problem-specific parameters, allowing for finding solutions faster and of better quality. 
The diffusion models used in our approach are trained on randomly cut solution trajectories. 
These trajectories are created by solving randomly generated problem instances with a kinodynamic motion planner. 
Experimental results show significant improvements up to 30 percent in both computation time and solution quality across varying robot dynamics such as second-order unicycle or car with trailer. 

\end{abstract}

\acrodef{adam}[ADAM]{Adaptive Moment Estimation}
\acrodef{asha}[ASHA]{Asynchronous Successive Halving}
\acrodef{cdf}[CDF]{Cumulative Distribution Function}
\acrodef{chomp}[CHOMP]{Covariant Hamiltonian Optimization for Motion Planning}
\acrodef{db-a*}[db-A$^*$]{Discontinuity Bounded A$^*$}
\acrodef{db-cbs}[db-CBS]{Discontinuity Bounded Conflict-Based Search}
\acrodef{ddpm}[DDPM]{Denoising Diffusion Probabilistic Model}
\acrodef{dmp}[DMP]{Dynamic Movement Primitive}
\acrodef{iqr}[IQR]{Interquartile Range}
\acrodef{komo}[KOMO]{k-Order Markov Path Optimization}
\acrodef{mlp}[MLP]{Multi Layer Perceptron}
\acrodef{mse}[MSE]{Mean Squared Error}
\acrodef{prm}[PRM]{Probabilistic Roadmap}
\acrodef{prodmp}[ProDMP]{Probabilistic Dynamic Movement Primitive}
\acrodef{promp}[ProMP]{Probabilistic Movement Primitive}
\acrodef{relu}[ReLU]{Rectified Linear Unit}
\acrodef{rl}[RL]{Reinforcement Learning}
\acrodef{rrt}[RRT]{Randomly Exploring Random Tree}

\section{Introduction}
\begin{figure}[ht]
	\centering
    	\includegraphics[width=0.9\linewidth]{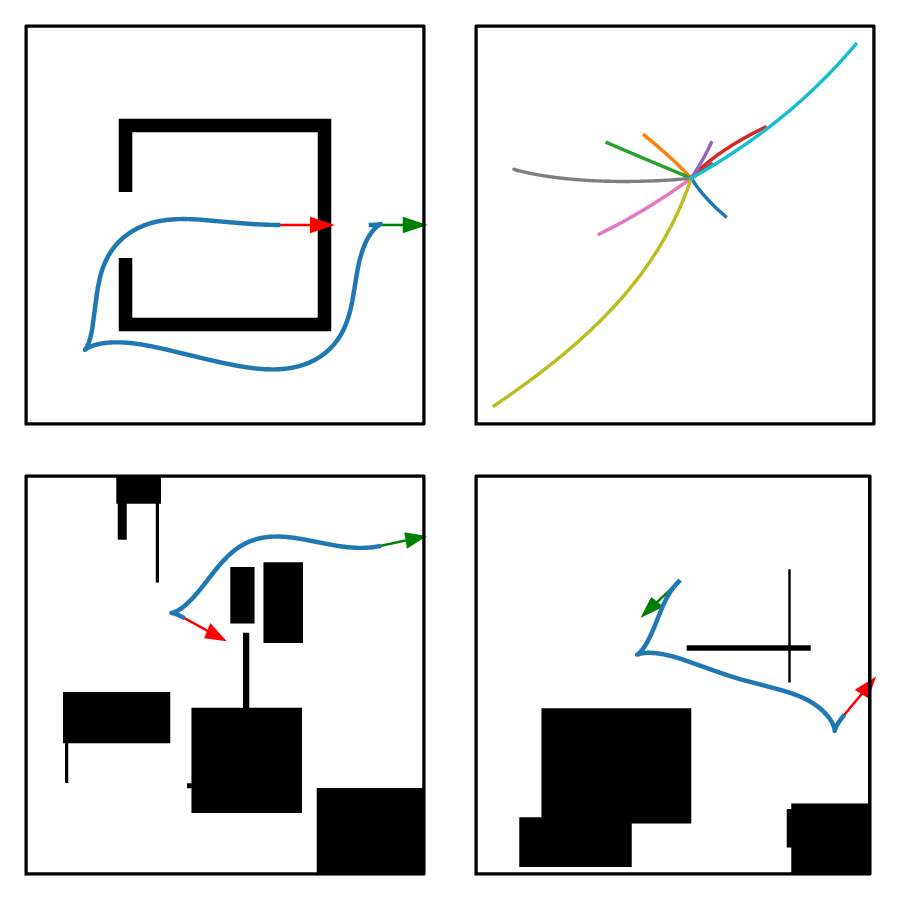}
	\caption[Image]{Examples for the $2^{\text{nd}}$ order unicycle - Top left: Bugtrap with solution (red: start, green: goal), Top right: Ten sampled motion primitives (starting from origin) Bottom row: Two random instances with a solution found by our method (red: start, green: goal)}\label{fig:overview}
\end{figure}

Kinodynamic motion planning \cite{donaldKinodynamicMotionPlanning1993} is a crucial part of robotics. 
It aims to find feasible motions that guide robots from a start state to a specified desired goal state while adhering to the robot's dynamic constraints, see \cref{fig:overview}. These trajectories can be composed from shorter subtrajectories, so-called motion primitives, which are short, pre-computed motions respecting the robot's dynamic constraints (see top left in \cref{fig:overview}).
It simultaneously optimizes an objective, such as time or energy consumption. Traditional motion planning methods include graph-based methods such as A$^*$ with motion primitives~\cite{hartFormalBasisHeuristic1968}, sampling-based approaches~\cite{LaValle1998RapidlyexploringRT}, or optimization-based methods~\cite{toussaintNewtonMethodsKorder2014}. 
While these approaches are effective in various scenarios, they come with limitations. 
For example, search-based methods require computing effective motion primitives, while optimization-based planners need an initial guess as a starting trajectory. 
Hybrid approaches are designed to overcome these limitations and have shown significant improvements in both computational efficiency and solution quality~\cite{honigDbDiscontinuityboundedSearchKinodynamic2022}.
A vital component of these hybrid approaches is the use of a set of motion primitives.
The selection of these motions to compose the set is a significant decision, which has shown to have a crucial impact on the computation time of the planning algorithm and the solution cost \cite{goddardSelectingMinimalMotion2023,poffald2014learning}. 
Current methods for the selection of motion primitives are random and independent of the problem instance, which may not fully utilize the planner's potential.

Diffusion models are a class of generative deep learning models that have achieved state-of-the-art performance in tasks like image~\cite{rombach2022high, rameshHierarchicalTextConditionalImage2022} and audio generation~\cite{kongdiffwave}.
In this paper, we introduce diffusion models to generate a set of efficient motion primitives for arbitrary problem instances. 
Furthermore, we condition the diffusion model on the characteristics of the problem instance to generate motion primitives specific to the problem.

We show that
\begin{itemize}
    \item the trained diffusion model can generate sets of valid motion primitives conditioned on a specific problem,
    \item our approach can generate sets of effective motion primitives for several robot dynamics and on diverse and difficult problem instances,
    \item the sets of motion primitives generated by our diffusion model result in a reduction of the solution cost and planning duration by $15\%$ for the first-order unicycle and by $30\%$ for the other evaluated robot dynamics compared to the baseline.
\end{itemize}

\section{Related Work}

Kinodynamic motion planning is a challenging problem due to the required reasoning over space and time over long time horizons.
Search-based methods using motion primitives can be applied to a variety of robot systems~\cite{likhachev2009}, including high-dimensional systems~\cite{dharmadhikari2020}.
Once motion primitives are computed, any variant of discrete path planning can be used without modifications. 
Sampling-based methods construct motions by sampling the state space and action space~\cite{sst}. 
Although solutions have probabilistic completeness guarantees, they are suboptimal and require post-processing to smooth the trajectory.
Optimization-based planners return a locally optimized solution, for example, by employing sequential convex programming (SCP)~\cite{schulman2014}. 
However, the success to compute a solution depends on an initial guess. 
In order to overcome the limitations of these planners, hybrid methods combining search and sampling, search and optimization, or sampling and optimization have been proposed. In this work, the hybrid planner db-A*~\cite{honigDbDiscontinuityboundedSearchKinodynamic2022} combining search, sampling, and optimization is used.

Recent works propose learning-based planners due to the ability of neural networks to recognize patterns in large datasets.

The combination of data-driven insights and model-based techniques can be applied to autonomous driving applications~\cite{pedrosaLearningMotionPrimitives2022}. 
Motion patterns in human driving data are generalized using the symmetry in vehicle dynamics. Here, a common dataset of recorded motions performed by human drivers is used to compute an optimal subset of selected primitives.
Our work also relies on a ground-truth dataset for training the diffusion model, but the main advantage of utilizing a generative model is that the synthesized primitives are similar but not always identical to the ground-truth. Thus, the model should be able to suggest a set of primitives potentially leading to a higher-quality solution.

Swift maneuvers of quadrotors can be achieved by building a motion primitive library using \ac{rl}~\cite{camciLearningMotionPrimitives2019}. 
The training data is comprised initially of recorded data from experiments in simulation and the real world. The \ac{rl} agent predicts a set of actions in the form of Bézier curves. The authors limit the action space to prevent predicted actions to cause undesireable states of the robot. In our proposed approach, the diffusion model also predicts consecutive actions but we utilize the robot dynamics to ensure feasible trajectories. RL agents can generalize to unseen states and environments but only relying on them for control might lead to unpredictable behavior of the robot if the agent fails to recover. Our method generates motion primitives obeying the robot dynamics constraints resulting in safe trajectories. 

\section{Problem Definition}

The state of the robot is defined as $\vq \in \mathcal Q \subset \mathbb R^{d_{q}}$, which is actuated by controlling the action $\vu \in \mathcal U \subset \mathbb R^{d_{u}}$. 
The workspace the robot operates in is given as $\sW \subseteq \mathbb R^{d_w}$ ($d_w\in\{2,3\}$). 
The collision-free space is $\sW_{\mathrm{free}} \subseteq \sW$.

We assume that a robot has the dynamics 
\begin{equation}
    \label{eq:dynamics}
    \dot \vq = \vf(\vq, \vu)\,.
\end{equation}
The Jacobian of $\vf$ with respect to $\vq$ and $\vu$ is assumed to be available in order to use gradient-based optimization. 

With zero-order hold discretization, \cref{eq:dynamics} can be framed as
\begin{equation}
    \label{eq:dynamics_discrete}
    \vq_{t+1} \approx \step(\vq_t, \vu_t) \equiv \vq_t + \vf(\vq_t, \vu_t)\Delta t,
\end{equation}
 where $\Delta t$ is sufficiently small to ensure that the Euler integration holds.

We use $\seqQ = \langle \vq_0, \vq_1, \ldots, \vq_T \rangle$ as a sequence of states of a robot sampled at times $0, \Delta t, \dots, T \Delta t$ and $\seqU = \langle \vu_0, \vu_1, \ldots, \vu_{T-1} \rangle$ as a sequence of actions applied to a robot for times $[0,\Delta t), [\Delta t, 2\Delta t), \ldots, [(T-1)\Delta t, T\Delta t)$.

Ultimately, the goal is to navigate the robot from its start state \(\vq_s\) to a goal state \(\vq_g\) as fast as possible with no collisions. This can be framed as the optimization problem with the objective of minimizing the arrival time of a robot:
\begin{align}
    \label{eq:opt1}
		  & \min_{\seqQ,\seqU,T} T \,,                                                          \\
		 &\text{\noindent s.t.}\begin{cases}
		\mathbf q_{t+1} = \step \left(\mathbf q_t, \mathbf u_t\right)             & \forall t \in \{0,\ldots,T-1\} \,, \\ 
        \vu_t \in \sU \,\,\,\, \vq_t \in \mathcal Q  & \forall t \in \{0,\ldots,T-1\} \,, \\ 
		  \mathcal B(\vq_t) \subset \mathcal W_\text{free}                 & \forall t \in \{0,\ldots,T\}  \,,     \\
		  \vq_0 = \vq_s; \,\, \vq_T = \vq_g,
		 \end{cases} \nonumber
\end{align}
where $\sB: \mathcal Q \to 2^\sW$ is a function that maps the configuration of a robot to a collision shape. 

Problem instances are characterized with $\vq_s$, $\vq_g$ and $\sW_{\text{free}}$, thus they will later be used to condition the diffusion model.
We train diffusion models to propose a set of subsequences of $\seqQ$ and $\seqU$ to the planner, such that \cref{eq:opt1} results in a lower value compared to using the planner with a conventionally generated motion primitive set.

\section{Background}
\subsection{Motion Primitives}
 
A motion primitive is a trajectory that fulfills the dynamics, control, and state
constraints. A single motion primitive is defined as a tuple $ \langle \seqQ, \seqU, T \rangle$, consisting of state sequences  $\seqQ = \langle \vq_0,..., \vq_T \rangle$ and control sequences $\seqU = \langle \vu_0, ..., \vu_{T-1} \rangle$, which obey the dynamics $\vq_{t+1} = \step(\vq_t, \vu_t)$. 
Motion primitives can be generated by solving two-point boundary value problems with random start and goal configurations in free space using nonlinear optimization~\cite{ortiz2024idb}. The resulting motions can be split into multiple pieces of different length (i.e., the number of states and controls). 

\subsection{Kinodynamic Motion Planning with db-A*} 
Kinodynamic Motion Planning with Discontinuity-Bounded A* (\textbf{kMP-db-A*}) is an iterative algorithm combining a search algorithm, discontinuity-bounded A* (db-A*), and trajectory optimization. The discrete planner db-A* uses motion primitives as graph edges and allows a user-defined discontinuity at the graph vertices. These discontinuities in the trajectory are repaired with trajectory optimization.
The pseudo-code of kMP-db-A* is shown in~\cref{alg:kmp-dbA*}. 
In every iteration, the following steps are performed: \emph{(i)} more motion primitives are added to the set $\mathcal M$ and the value of the discontinuity bound $\delta$ is decreased (\cref{alg:overview:sM} - \cref{alg:overview:delta}); \emph{(ii)} the discrete planner db-A* computes a trajectory using the current set of motion primitives. The computed trajectory may contain discontinuous jumps up to $\delta$ causing dynamic constraint violations (\cref{alg:overview:dbAstar}); \emph{(iii)} the discontinuous jumps are repaired using optimization, which uses the result of db-A* as an initial guess (\cref{alg:overview:opt}); \emph{(iv)} More motion primitives are extracted from the optimization output (\cref{alg:overview:extract}). 

Discontinuity-Bounded A* (\textbf{db-A*}) is an extension of the well-known A$^*$ algorithm~\cite{honigDbDiscontinuityboundedSearchKinodynamic2022}. It searches a graph of motion primitives.
These motion primitives are used as graph edges to connect states, representing graph nodes, with user-configurable discontinuity $\delta$.
Db-A* is an informed search which explores nodes based on $f(\vq)=g(\vq)+h(\vq)$, where $g(\vq)$ is the cost-to-come. The node with the lowest $f$-value is expanded using the collision-free motion primitives. The output of db-A* is a $\delta$-discontinuity-bounded solution. 

\begin{algorithm}[t]
	\caption{kMP-db-A* -- Kinodynamic Motion Planning with db-A*\cite{honigDbDiscontinuityboundedSearchKinodynamic2022}}
    \label{alg:kmp-dbA*}
    \tcc{Changes compared to the original kMP-dbA* are in blue}
	\KwData{$\vq_s, \vq_g, \sW_{\mathrm{free}} $}
	\KwResult{$\seqQ, \seqU$ }
	\label{alg:overview}
	\DontPrintSemicolon
	\SetKwFunction{AddPrimitives}{AddPrimitives}
	\SetKwFunction{GeneratePrimitives}{GeneratePrimitives}
	\SetKwFunction{ExtractPrimitives}{ExtractPrimitives}
	\SetKwFunction{ComputeDelta}{ComputeDelta}
	\SetKwFunction{ChooseDelta}{DecreaseDelta}
	\SetKwFunction{ChoosePrimitives}{IncreasePrimitives}
	\SetKwFunction{DiscontinuityBoundedAstar}{Db-A*}
	\SetKwFunction{Optimization}{Optimization}
	\SetKwFunction{Report}{Report}
	$\sM_0 \leftarrow \emptyset$ \Comment*{Initial Set of motion primitives}
	$c_{\mathrm{max}} \leftarrow \infty$ \Comment*{Solution cost bound}
	\For{$i=1,2,\ldots$}{
		$\sM_i \leftarrow \sM_{i-1} \cup \tikzmk{A} \{ \GeneratePrimitives(i, l_n) \}_{n=1}^N$ \markbluelineHalf{-5}\label{alg:overview:sM}\;
		$\delta_i \leftarrow \ChooseDelta(i)$\label{alg:overview:delta}\;
		$\seqQ_d, \seqU_d  \leftarrow$ \DiscontinuityBoundedAstar{$\vq_s, \vq_g, \sW_{\mathrm{free}}, \sM_i, \delta_i, c_{\mathrm{max}}$}\label{alg:overview:dbAstar}\;
		\If{$\seqQ_d, \seqU_d$ successfully computed}{
			$\seqQ, \seqU \leftarrow$ \Optimization{$\seqQ_d, \seqU_d, \vq_s, \vq_g, \sW_{\mathrm{free}} $}\label{alg:overview:opt}\;
			\If{$\seqQ, \seqU$ successfully computed}{
				\Report{$\seqQ, \seqU$}\label{alg:overview:report} \Comment*{New solution found}
				$c_{\mathrm{max}} \leftarrow \min(c_{\mathrm{max}}, J(\seqQ, \seqU))$ \label{alg:overview:cmax} \Comment*{Cost bound}
			}
			$\sM_{i} \leftarrow  \sM_i \cup    \ExtractPrimitives(\seqQ, \seqU)\label{alg:overview:extract}$\;
		}
	}
\end{algorithm}

\subsection{Diffusion Models}\label{subsec:diffusion}
\textbf{Diffusion Models}~\cite{sohl2015deep}, specifically \acfp{ddpm} \cite{ho2020denoising} (see also \cite[chap. 20]{Bishop2023}), are generative deep learning models generating synthetic data from noise. %

The process of learning and inference, however, differs from regular deep learning models. The forward process is defined as a Markov chain~\cite{meynMarkovChainsStochastic2009}, in which Gaussian noise is added to the original input data over $T$ discrete time steps. The goal is to learn the reverse process, removing noise from the corrupted data to reconstruct the original input. Given the original input $\mathbf{p}_0 \in\mathbb{R}^d$ of dimension $d$, the forward process generates a sequence of $T$ noisy datapoints $\mathbf{p}_1, \mathbf{p}_2, \ldots, \mathbf{p}_T$ by adding random Gaussian noise $\epsilon_t \sim \mathcal{N}( \epsilon_t | \mz, \mI)$, with $\mI$ as the unit matrix, and noise schedule $\alpha_t$:

\begin{equation}
    \mathbf{p}_t = \sqrt{\alpha_t}\mathbf{p}_0+\sqrt{1- \alpha_t}\mathbf\epsilon_t\label{eq:ddpm_noise}
\end{equation}

The noise schedule can in general be defined as $\alpha_t = \prod_{\tau=1}^t(1-\beta_{\tau})$, with $\beta_t\in[0,1]$. It ensures $\mathbb E(\mathbf{p}_{t+1})\leq\mathbb E(\mathbf{p}_{t})$ and $\text{Var}(\mathbf{p}_{t+1}) \leq \text{Var}(\mathbf{p}_{t})$. There are a few variants for different noise schedules, with the linear schedule as the most basic one~\cite{chenImportanceNoiseScheduling2023}.

The reverse process aims to \textit{denoise} $\mathbf{p}_t$ to obtain $\mathbf{p}_0$. Equation \cref{eq:ddpm_noise} can be rearranged for this purpose as:
\begin{align}
    \mathbf{p}_0 = \frac{1}{\sqrt{\alpha_t}} \mathbf{p}_t - \frac{\sqrt{1 - \alpha_t}}{\sqrt{\alpha_t}}\epsilon_t
\end{align}
\citet{ho2020denoising} found that a neural network $g$ with parameters $\phi$ can be used to predict the total noise added to $\mathbf{p}_0$ to obtain $\mathbf{p}_T$ trained with the following objective function:
\begin{align}\label{eq:diff_loss}
    \mathcal{L}_\phi = - \sum_{t=1}^T \|g_\phi(\sqrt{\alpha_t}\mathbf{p}_0 + \sqrt{1-\alpha_t}\epsilon_t, t) - \epsilon_t \|_2 \;.
\end{align}
The model architecture of $g$ for a basic diffusion model can be equivalent to a \ac{mlp}.

The inference, or \textit{sampling}, to generate a new, synthetic datapoint starting from a sample of the Gaussian distribution $p(\mathbf{p}_T)$, $g$ can be leveraged to define the function $h$:
\begin{align}
    h(\mathbf{p}_t, t) = \frac{1}{\sqrt{1 - \beta_t}}\left(\mathbf{p}_t - \frac{\beta_t}{\sqrt{1-\alpha_t}}g_\theta(\mathbf{p}_t, t)\right)
\end{align}
Each intermediate datapoint $[\mathbf{p}_T,\ldots, \mathbf{p}_1]$ can therefore be predicted with:
\begin{align}
    \mathbf{p}_{t-1} = h(\mathbf{p}_t, t) + \sqrt{\beta_t}\epsilon,
\end{align}
for all $t\in [T, \ldots, 2]$, where $\epsilon\sim\mathcal{N}(\epsilon|\mz,\mI)$.
Finally, we can retrieve $\mathbf{p}_0 = h(\mathbf{p}_1, 1)$.

With this basic approach, the model can only sample from \textbf{one} distribution. In many cases, the desired distribution can change depending on some conditions. In the use case of generating images, this could, for example, be the content or the shape of an object in the image. To differentiate between those different distributions, new values, called conditioning $p_\text{condition}\in\mathbb R^c$, are added to the model's input \cite{zhang2023shiftddpms, dieleman2022guidance}.

When utilizing conditioning in a diffusion model, the conditioning in the training data has to be distributed over the entire spectrum that will later be sampled. If this is not fulfilled, it could lead to undesirable effects like hallucinations. A model is hallucinating when its output is incorrect or nonsensical \cite{maleki2024ai, dewynterEvaluationLargeLanguage2023}.

\section{Approach}\label{sec:approach}
In this section, we decribe the methods to train and deploy the diffusion model for generating motion primitives that can generalize to arbitrary problem instances.

To train the model, first a dataset of valid, ground-truth motion primitives is necessary. Therefore, we utilize kMP-db-A$^*$ with a conventional, randomly generated motion primitive dataset, to generate this training dataset on random problem instances.

\subsection{Dataset Creation}
\label{subsec:random_instances}
To ensure that the diffusion model generalizes to a variety of problem instances, a large amount of instances differing in difficulty and environment layouts is necessary. Creating them by hand is tedious and prone to introducing a bias in the instances themselves. Therefore, an automated approach to generating random problem instances is favorable. 

We, therefore, generate random instances by placing rectangular obstacles with random dimensions at a random position in the workspace. These obstacles may overlap to allow complex, non-convex scenarios.
The obstacle density, i.e., the percentage of obstacles occluding the workspace, is user-specified. %
The start and goal configurations are placed randomly in the workspace, asserting that the robot is not in collision with the environment.

We then use kMP-db-A$^*$ to solve a set number of randomly generated problems with randomly generated motion primitives. This process is repeated several times to ensure that the random choice of motion primitives has no major influence on the general distribution of the solutions. A single query of kMP-db-A$^*$ can yield zero, one or multiple individual solutions, depending on the number of \texttt{Report} calls (see \cref{alg:kmp-dbA*}, \cref{alg:overview:report}). Each individual solution consists of action and state sequences. The solutions are then randomly cut into motion primitives of differing lengths and extracted together with the duration, the cost and the condition variables described in \cref{subsec:conditioning}. This information is stored in a dataset. We create $N$ separate subsets, one for each motion primitive length.

\subsection{Conditioning}
\label{subsec:conditioning}
Utilizing conditioning plays an important role in guiding the model to generate motion primitives that are well-suited to specific problem instances. To this end, the models used in this work are conditioned on variables that capture both solution-specific and problem-instance-specific characteristics. 

There are two solution-specific variables. The first one is the relative cost, which is defined as the cost of the solution divided by the best solution achieved for the same instance. The second one is the relative location, which is defined as the position of a given motion primitive within the solution, normalized between $0$ and $1$, and hence providing a sense of progress.

The problem-instance-specific variables that are used are the width and height of the environment, the obstacle density, as well as the non-translational parts of the start and goal configuration.

These condition variables had the most positive impact on the diffusion model's performance. For a detailed ablation study of the different conditioning options see \cite{masterthesisjulius}.

\subsection{Training}
For each robot dynamic, we train $N$ diffusion models. Each of the $n \in N$ models are trained to only reproduce the motion primitives of a particular length, i.e., replicate one of the $n \in N$ subsets in the ground-truth dataset. This allows us to synthesize motion primitives of different lengths by simply querying each of these diffusion models to generate a subset for the final set of motion primitives. The decision on how the different lengths are distributed in the final set, is based on tunable hyperparameters.

The output of each model consists of a starting state and all consecutive actions. The missing intermediate states will later be reconstructed via the robot dynamics during inference. If the model were to output these states directly, it would need to implicitly learn the robot dynamics, which lead to worse results (see also \cite{masterthesisjulius} for an ablation study).
Depending on the dynamics of the chosen robot, some parts of the starting state can be dropped as well. For example, given translation invariant dynamics, the translational part of the starting state is not necessary.

As the goal of the training is to reduce the error between the prediction of the added noise and the actual added noise as formulated in \cref{eq:diff_loss} and some actions for certain dynamics include angles, the representation of the angles has to be changed from the standard scalar representation. When using an error function like \cref{eq:diff_loss}, the error will be large if $(-\pi,\pi)=2\pi^2$, even though they represent the same angle. Therefore, the angles are represented as $[\sin\theta, \cos\theta]^T$, resulting in a low error for any two angles of the same absolute value.

\subsection{Deployment}
The size of the motion primitives set $\mathcal M$ in \cref{alg:kmp-dbA*} is a user-defined parameter, changing with iteration $i$. We denote this parameter as $L_i$. 
Given the value $L_i$, we sample from the $N$ models in a consecutive order to generate motion primitives of certain size $l_n=p_n L_i$ (\cref{alg:overview:sM}). This value is determined via the tunable hyperparameters $p_n$, denoting the percentage of $L_i$ that should be generated by each individual model. To avoid inference during the execution of kMP-db-A$^*$, a motion primitive set of size $5L_i$ is generated and cached pre-execution.

\section{Results}
\label{sec:results}

We consider robots with different dynamics: unicycle, $2^{\text{nd}}$-order unicycle and car with trailer, see~\cite{honigDbDiscontinuityboundedSearchKinodynamic2022} for dynamics and bounds.

For testing scenarios, we focus on cases where the workspace has varying dimensions and obstacles.
Some problem instances are from~\cite{honigDbDiscontinuityboundedSearchKinodynamic2022} to test canonical cases from the literature.

In each environment we test kMP-db-A* using randomly generated motion primitives (\textit{baseline}) with kMP-db-A* using motion primitives generated by diffusion models conditioned on the environment (\textit{model}). 
We analyze success rate $(p)$, computational time until the first solution is found $(d)$, and cost of the first solution $(c^\textrm{first})$, and the best solution cost $(c^{\textrm{best}})$. Note that the cost is a time equal to the control duration over the path. The duration includes only the main loop of kMP-dbA$^*$, excluding inference time of the diffusion model, as well as loading of any dataset. In the following results, the duration and costs are presented as \textbf{notion of regret} $r$, which is defined as $r_x = 100\frac{x-\tilde{x}_\textrm{Baseline}}{\tilde{x}_\textrm{Baseline}}$, 
with $x\in\{d, c^\textrm{first}, c^\textrm{best}\}$ and $\tilde{x}$ being the median of $x$. Negative regret for both duration and cost, therefore, indicates an improvement over the baseline. An instance is not solved successfully if an incorrect solution is returned or no solution is found after the time limit. A trial which did not succeed does not have a cost or duration attached to it and is therefore excluded from the calculation of the regret. 

For kMP-db-A* we use the existing implementation from~\cite{honigDbDiscontinuityboundedSearchKinodynamic2022}.
The diffusion model training script is written in Python using PyTorch ~\cite{paszke2019pytorch} and implemented according to \cref{subsec:diffusion}. Each model utilizes fully connected layers with \ac{relu} as the activation function. The gradient descent steps are perfomed by the \ac{adam} optimizer. The size and number of hidden layers, the learning rate, the number of denoising steps, and the noise schedule alongside the conditioning mentioned in \cref{subsec:conditioning} are used as tunable hyperparameters. Instead of relying on the validation loss as a metric to verify the models performance, kMP-db-A$^*$ is queried every $10$ epochs as a benchmark. The success, cost and duration of these benchmarks are then combined in a single metric, which has a large weight on the success. 
The hyperparameters are tuned using the search algorithm Optuna ~\cite{DBLP:conf/kdd/AkibaSYOK19} in conjunction with the \ac{asha} scheduler~\cite{MLSYS2020_a06f20b3}.
For each of the dynamics tested, the hyperparameters are tuned for $100$ randomly initiated trials with a maximum of $300$ epochs. The best model is then trained for $1000$ epochs and used for the benchmarking. We use a workstation with an AMD Ryzen 7 5800X @ 3.8 GHz, 32 GB RAM, and Ubuntu 22.04.

\begin{table}[t]
\caption{Benchmark Results for 10 random instances with 20 trials each for varying initial motion primitives, bold entries are the best for the dynamics}
\centering
\setlength{\tabcolsep}{5pt}
\begin{tabular}{l|r||rr|rrr}
\toprule
 Dynamics&$\textrm{primitives}_0$&  $p_\textrm{B}$&  $p_\textrm{M}$ & $\Tilde{r}_{d}$ &  $\Tilde{r}_c^\mathrm{first}$    & $\Tilde{r}_c^\mathrm{best}$ \\
\midrule
 \multirow{4}{*}{\shortstack{$\text{1}^{\text{st}}$-order\\ Unicycle}}&100&\textbf{100.0} &\textbf{100.0} &\textbf{-16.9} &\textbf{-17.2} &\textbf{-17.2} \\
  &150& \textbf{100.0} &\textbf{100.0} & -11.1 &-14.7 &-14.7\\
  &200& \textbf{100.0} & \textbf{100.0} & -10.8 & -16.7 & -16.7\\
  &250& \textbf{100.0} & \textbf{100.0} & -10.6 & -13.0 & -13.0\\
\midrule
 \multirow{4}{*}{\shortstack{$\text{2}^{\text{nd}}$-order\\ Unicycle}}&100&96.5 &98.5 &-29.9 &\textbf{-34.0} &\textbf{-34.0} \\
  &150& 98.5 &\textbf{99.0} & -56.3 &-21.7 &-21.5\\
  &200& \textbf{99.0} & 98.5 & -69.6 & -8.2 & -19.4\\
  &250& \textbf{99.0} & \textbf{99.0} & \textbf{-75.0} & -5.4 & -13.4\\
\midrule
 \multirow{4}{*}{\shortstack{Car with\\ trailer}}&100&99.0 &\textbf{100.0} &-37.7 &\textbf{-28.6} &-27.7 \\
  &150& 98.0 & 98.5 & \textbf{-42.6} &-28.1 &\textbf{-28.0}\\
  &200& 99.0 & 98.0 & -37.7 & \textbf{-28.6} & -27.3\\
  &250& 97.0 & 96.5 & -20.3 & -24.7 & -24.2\\
  
\bottomrule
\end{tabular}
\label{tab:stats4}
\end{table}

\subsection{Benchmarking}
\begin{figure*}[tb]
	\centering
	\includegraphics[width=1\textwidth]{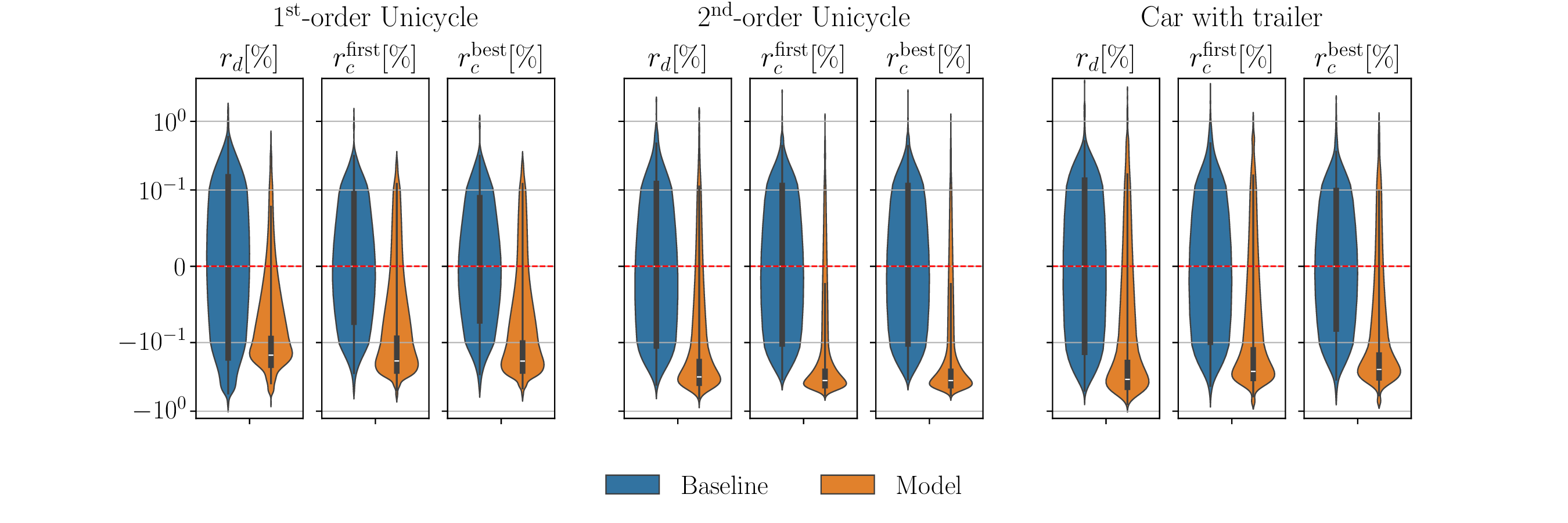}
	\caption{Violin plots of regret for duration and costs for three dynamics. The introduced diffusion model outperforms the baseline across different robot dynamics and in all metrics.}
	\label{fig:benchmarks}
\end{figure*}

The first benchmark runs kMP-db-A$^*$ with $100$ initial motion primitives and $\delta_0=0.5$ for the unicycles and $\delta_0=0.9$ for the car with trailer. The maximum allowed time is defined as \SI{1.5}{s}. The duration and costs are plotted in \cref{fig:benchmarks} for the different dynamics respectively. The success rates for all models as well as for the baseline are $>98\%$. The models outperform their respective baselines over all metrics. The model for the $2^{\text{nd}}$-order unicycle performs the best, with over $30\%$ improvements in all metrics compared to the baseline. 

For comparative purposes, a model without any conditioning for the $1^\text{st}$-order unicycle was tested as well. The success rate of this model performed equivalent to that of the model with conditioning. For duration and cost, however, it performed worse. The median improvement in duration is $2\%$ faster than the baseline, while the model with conditioning achieved a median improvement of $15\%$. Both cost metrics showed an improvement of $10\%$ while the model with conditioning was able to improve upon the baseline by $18\%$. 
\subsection{Ablation studies}
The amount of initial motion primitives (primitives$_0$), as well as the starting discontinuity $\delta_0$ are important parameters to tune for kMP-db-A$^*$. For this reason, the benchmark is repeated with a varying number of initial motion primitives (see~\cref{tab:stats4}) or $\delta$ (see~\cref{tab:stats5}). 

When adapting the initial amount of motion primitives, the success rates of all dynamics is consistent with the baseline. The duration and costs show an improvement over all trials. The costs show a slightly lower improvement with increasing starting primitives. The relative duration improvement for the $2^{\text{nd}}$-order unicycle is scaling particularly well with an increased amount, while the other dynamics do not show the same benefit.

Changing $\delta_0$ does not show any impact for the unicycle as well as the car with trailer. The $2^{\text{nd}}$-order unicycle, however, outperforms the baseline success-rate for $\delta_0=0.3$ with the baseline solving only $23\%$ of the trials, while the model is still able to solve $79\%$ of the trials. The relative cost improvement is high as well, although it is not necessarily representative given the low success rate of the baseline.

The performance of the models is also evaluated for some canonical cases from the literature. The results are shown in \autoref{tab:stats6}. These instances are not included in the training data in any way. The random instances the models are trained on do not show the same structure as the canonical, handcrafted instances. Still, the models outperform the baseline in all metrics. Unlike the previous tests, the improvements in cost are now generally larger than the improvements in duration. 

\begin{table}[t]
\caption{Benchmark Results for 10 random instances with 20 Trials each for varying $\delta_0$, bold entries are the best for the dynamics}
\centering
\setlength{\tabcolsep}{5pt}
\begin{tabular}{l|r||rr|rrr}
\toprule
 Dynamics&$\delta_0$&  $p_\textrm{B}$&  $p_\textrm{M}$ & $\Tilde{r}_{d}$ &  $\Tilde{r}_c^\mathrm{first}$    & $\Tilde{r}_c^\mathrm{best}$ \\
\midrule
 \multirow{3}{*}{$\text{1}^{\text{st}}$-order Unicycle}&$0.3$&\textbf{100.0} &\textbf{100.0} &-1.3 &-15.8 &-15.8 \\
  &$0.5$& \textbf{100.0} &\textbf{100.0} & \textbf{-12.3} &\textbf{-20.8} &\textbf{-20.8}\\
  &$0.7$& \textbf{100.0} & \textbf{100.0} & -5.8 & -18.7 & -18.7\\
\midrule
 \multirow{3}{*}{$\text{2}^{\text{nd}}$-order Unicycle}&$0.3$&23.5 &79.0 &\textbf{-25.7} &\textbf{-56.5} &\textbf{-56.5} \\
  &$0.5$& 97.0 &96.5 & -23.6 &-37.5 &-37.5\\
  &$0.7$& \textbf{100.0} & \textbf{100.0} & -18.6 & -26.5 & -26.5\\
\midrule
 \multirow{3}{*}{Car with trailer}&$0.7$&\textbf{100.0} &93.5 &-40.6 &-30.6 &\textbf{-33.5} \\
  &$0.9$& 99.5 &99.0 & -36.7&\textbf{-35.0} &-31.8\\
  &$1.1$& 43.0 & 41.5 & \textbf{-45.8} & -24.8 & -10.8\\
  
\bottomrule
\end{tabular}
\label{tab:stats5}
\end{table}

\begin{table}[t]
\caption{Benchmark Results for selected canonical problem instances, 20 trials each, bold success-rate entries are the best for the dynamics }
\centering
\setlength{\tabcolsep}{5pt}

\begin{tabular}{l|r||cc|rrr}
\toprule
 Dynamics&Instance&  $p_\textrm{B}$&  $p_\textrm{M}$ & $\Tilde{r}_{d}$ &  $\Tilde{r}_c^\mathrm{first}$    & $\Tilde{r}_c^\mathrm{best}$ \\
\midrule
\vtop{\hbox{\strut $\text{1}^{\text{st}}$-order}\hbox{\strut Unicycle}}&Bugtrap&\textbf{100.0} &\textbf{100.0} &-10.6 &-20.7 &-20.7 \\
\midrule
 \multirow{2}{*}{\vtop{\hbox{\strut $\text{2}^{\text{nd}}$-order}\hbox{\strut Unicycle}}}&Bugtrap&\textbf{100.0} &\textbf{100.0} &-2.45 &-34.8 &-34.8 \\
  &Park& \textbf{100.0} &\textbf{100.0} & -13.4 & -32.1& -32.1\\
\midrule
 \multirow{2}{*}{\vtop{\hbox{\strut Car with}\hbox{\strut trailer}}}&Kink&\textbf{100.0} &\textbf{100.0} &-31.1 &-8.48 &-12.3 \\
  &Park& \textbf{100.0} & \textbf{100.0} & -49.1 & -20.4 & -20.4\\
  
\bottomrule
\end{tabular}
\label{tab:stats6}
\end{table}

\section{Conclusion}
In this paper, we present a new approach for generating sets of motion primitives for kinodynamic motion planning using diffusion models. These models incorporate problem-specific parameters to generate datasets adapted to each problem instance, improving both efficiency and solution quality. The results demonstrate that our approach reduces the planning computation time and solution cost compared to the baseline method. 
The model performs particularly well for the second-order unicycle and car with trailer, where the solution cost and planning duration reduction exceeds $30\%$, highlighting the effectiveness of diffusion models in improving motion planning performance.

In future work, we aim to extend our work by additional dynamics, such as multirotors. The current conditioning on the environment is limited to statistical information and can be expanded to include a representation of the workspace.
\printbibliography

\end{document}